\pdfoutput=1

\documentclass[11pt]{article}

\usepackage{acl}
\usepackage{times}
\usepackage{latexsym}

\usepackage[T1
]{fontenc}

\usepackage[utf8]{inputenc}

\usepackage{booktabs}
\usepackage{multirow}
\usepackage{xspace}
\usepackage{amsmath}
\usepackage{amssymb}
\usepackage{soul}

\usepackage{subcaption}
\usepackage{microtype}
\usepackage{graphicx}
\usepackage{lipsum}
\usepackage[most]{tcolorbox}
\newcommand{\fon}[1]{\fontfamily{#1}\selectfont}
\usepackage{colortbl}

\usepackage{pifont}

\definecolor{darkgreen}{HTML}{548235}
\definecolor{darkred}{HTML}{C00000}
\definecolor{darkerblue}{HTML}{240394}
\definecolor{darkblue}{HTML}{2e75B6}
\definecolor{darkyellow}{HTML}{BF9000}
\definecolor{darkpurple}{HTML}{7030A0}
\definecolor{lightgray}{HTML}{e0e0e0}
\definecolor{lg}{HTML}{e6fce6}
\definecolor{ly}{HTML}{ffffeb}
\definecolor{lb}{HTML}{e3f0fa}
\definecolor{lp}{HTML}{fae3e3}

\lstdefinelanguage{textgame}{
  keywords={ OBJECT_RULES, ACTION_RULES, SCORE_RULES, GAME_STATE, GAME_STATE_DIFFERENCE, GAME_OBSERVATION, OBJECT_CLASS_CODE, GAME_CODE},
  keywordstyle=\color{blue}\bfseries,
  sensitive=true,
  breaklines=true,
  columns=fullflexible,
  basewidth = {.6em},
  breakindent = {0em},
  tabsize=1,
  aboveskip=0em,
  belowskip=0em,
  comment=[l]{>},
  commentstyle=\color{purple}\ttfamily,
  stringstyle=\color{blue}\ttfamily
}

\newcommand{\methodname}{LLM-as-a-Simulator\xspace}
\newcommand{\methodnameshort}{LLM-Sim\xspace}
\newcommand{\dataname}{\textsc{ByteSized32}-State-Prediction\xspace}
\newcommand{\datanameshort}{\textsc{ByteSized32-SP}\xspace}
\newcommand{\statechanged}{\textit{dynamic}\xspace}
\newcommand{\stateunchanged}{\textit{static}\xspace}
\newcommand{\Tick}{\text{Environment-driven}\xspace}
\newcommand{\tick}{\text{environment-driven}\xspace}
\newcommand{\tickshort}{\text{env}\xspace}

\newcommand{\act}{\text{action-driven}\xspace}
\newcommand{\Act}{\text{Action-driven}\xspace}
\newcommand{\actshort}{\text{act}\xspace}

\newcommand{\eg}{e.g.,\xspace}

\newcommand{\code}[1]{\texttt{#1}\xspace}

\title{Can Language Models Serve as Text-Based World Simulators?}

\author{Ruoyao Wang$^{\dagger}$, Graham Todd$^{\ddagger}$, Ziang Xiao$^{\spadesuit}$, Xingdi Yuan$^{\diamondsuit}$  \\ {\bf Marc-Alexandre Côté$^{\diamondsuit}$}, {\bf Peter Clark$^{\clubsuit}$},  {\bf Peter Jansen$^{\dagger\clubsuit}$} \\
$^{\dagger}$University of Arizona  ~~~~~ $^{\diamondsuit}$Microsoft Research Montréal \\
$^{\ddagger}$New York University ~~~~~ $^{\spadesuit}$Johns Hopkins University ~~~~~ $^{\clubsuit}$Allen Institute for AI \\
\texttt{\{ruoyaowang,pajansen\}@arizona.edu} ~~~~~ \texttt{gdrtodd@nyu.edu} \\ \texttt{ziang.xiao@jhu.edu} ~~~~~ \texttt{\{eric.yuan,macote\}@microsoft.com} \\ \texttt{PeterC@allenai.org}
}

\begin{document}
\maketitle
\begin{abstract}
Virtual environments play a key role in benchmarking advances in complex planning and decision-making tasks but are expensive and complicated to build by hand. Can current language models themselves serve as world simulators, correctly predicting how actions change different world states, thus bypassing the need for extensive manual coding? Our goal is to answer this question in the context of text-based simulators. Our approach is to build and use a new benchmark, called \dataname, containing a dataset of text game state transitions and accompanying game tasks. We use this to directly quantify, for the first time, how well LLMs can serve as text-based world simulators. We test GPT-4 on this dataset and find that, despite its impressive performance, it is still an unreliable world simulator without further innovations. This work thus contributes both new insights into current LLM's capabilities and weaknesses, as well as a novel benchmark to track future progress as new models appear.

\end{abstract}

\section{Introduction and Related Work}
\label{section:intro}


\begin{figure}[t!]
\centering
\includegraphics[width=\linewidth]{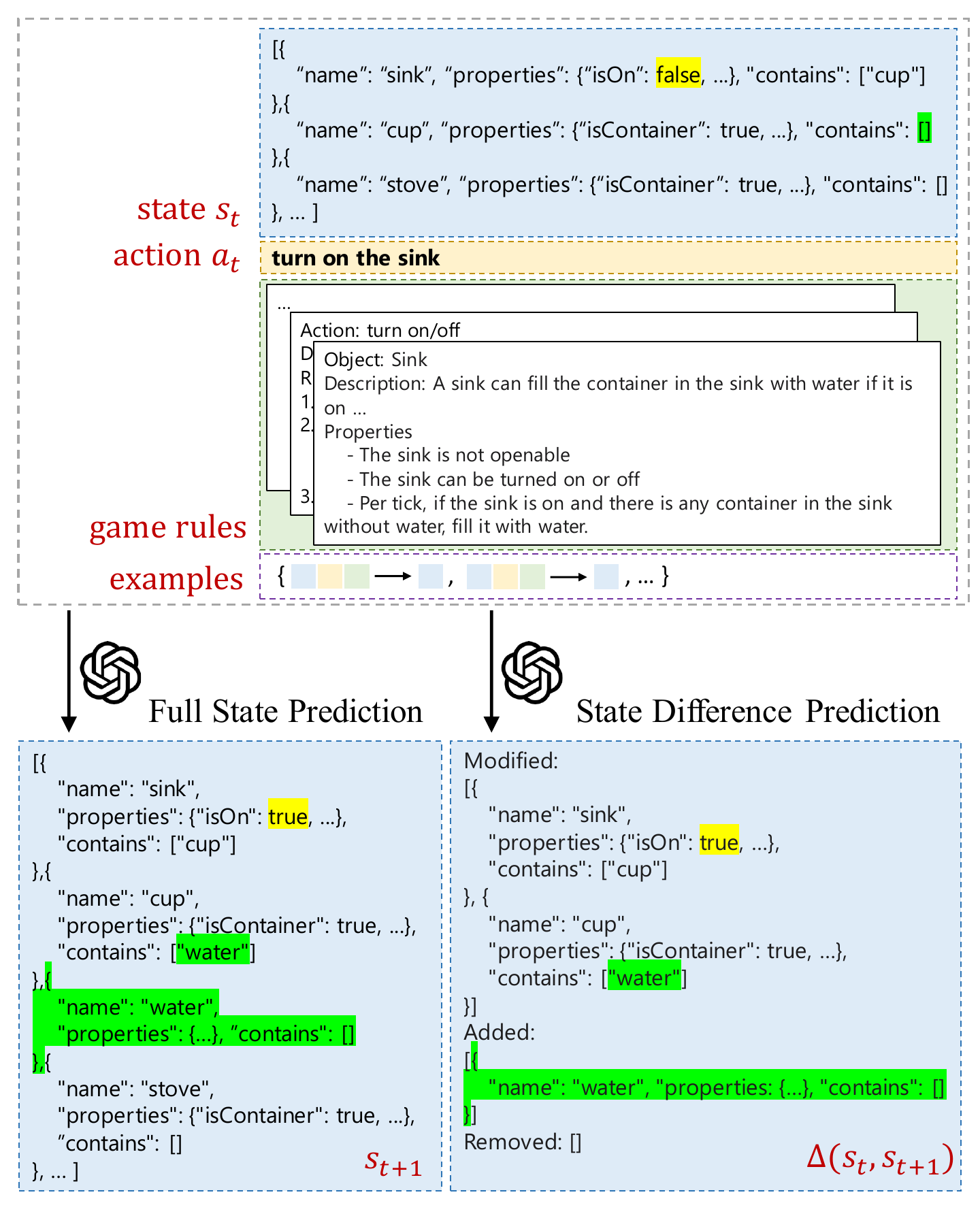}
    \caption{An overview of our two approaches using an LLM as a text game simulator. The example shows the process that a cup in the sink is filled by water after turning on the sink. The full state prediction includes all objects in the game including the unrelated stove, while the state difference prediction excludes the unrelated stove. State changes caused by $\mathcal{F}_{\actshort}$ and $\mathcal{F}_{\tickshort}$ are highlighted in \colorbox{yellow}{yellow} and \colorbox{green}{green}, respectively.} 
    \label{fig:intro}
    \vspace{-2em}
\end{figure}

Simulating the world is crucial for studying and understanding it. 
In many cases, however, the breadth and depth of available simulations are limited by the fact that their implementation requires extensive work from a team of human experts over weeks or months. Recent advances in large language models (LLMs) have pointed towards an alternate approach by leveraging the huge amount of knowledge contained in their pre-training datasets. But are they ready to be used directly as simulators?

We examine this question in the domain of text-based games, which naturally express the environment and its dynamics
in natural language and have long been used as part of advances in decision making processes \cite{cote18textworld, Fan_Urbanek_Ringshia_Dinan_Qian_Karamcheti_Prabhumoye_Kiela_Rocktaschel_Szlam_Weston_2020, urbanek2019learning, shridhar2020alfworld, hausknecht2020interactive, jansen2022systematic, wang-etal-2023-bytesized32}, information extraction \cite{ammanabrolu2020graph, adhikari2020learning}, and artificial reasoning \cite{wang2022scienceworld}.

Broadly speaking, there are two ways to leverage LLMs in the context of world modeling and simulation. The first is \textit{neurosymbolic}: a number of efforts use language models to generate code in a symbolic representation that allows for formal planning or inference \cite{liu2023llm+, nottingham2023embodied, wong2023word,tang2024worldcoder}. \textsc{Reasoning via Planning (RAP)} \cite{hao2023reasoning} is one such approach -- it constructs a world model using LLM priors and then uses a dedicated planning algorithm to decide on agent policies (LLMs themselves continue to struggle to act directly as planners \cite{valmeekam2023planning}). Similarly, \textsc{ByteSized32} \cite{wang-etal-2023-bytesized32} tasks LLMs with instantiating simulations of scientific reasoning concepts in the form of large \textsc{Python} programs. These efforts are in contrast to the second, and comparatively less studied, approach of \textit{direct simulation}. For instance, \textsc{AI-Dungeon} represents a game world purely through the generated output of a language model, with inconsistent results \cite{aidungeon}. In this work, we provide the first quantitative analysis of the abilities of LLMs to directly simulate virtual environments. We make use of \textit{structured representations} in the \textsc{JSON} schema as a scaffold that both improves simulation accuracy and allows for us to directly probe the LLM's abilities across a variety of conditions.


In a systematic analysis of GPT-4~\cite{achiam2023gpt}, we find that LLMs broadly fail to capture state transitions not directly related to agent actions, as well as transitions that require arithmetic, common-sense, or scientific reasoning. Across a variety of conditions, model accuracy does not exceed 59.9\% for transitions in which a non-trivial change in the world state occurs. These results suggest that, while promising and useful for downstream tasks, LLMs are not yet ready to act as reliable world simulators without further innovation.\footnote{Code and data are available at \url{https://github.com/cognitiveailab/GPT-simulator}.}

\section{Methodology}
\label{section:method}

\label{section:notations}
We examine the abilities of LLMs to serve as world simulators in text-based virtual environments, in which an agent receives observations and proposes actions in natural language in order to complete certain objectives. Each text environment can be formally represented as a goal-conditioned partially observable Markov decision process (POMDP)~\citep{kaelbling1998planning} with the 7-tuple $(S, A, \mathcal{T}, O, R, C, D)$, where $S$ denotes the state space, $A$ denotes the action space, $\mathcal{T}: S\times A \rightarrow S$ denotes the transition function, $O$ denotes the observation function, $R:S\times A \rightarrow \mathbb{R}$ denotes the reward function, $C$ denotes a natural language ``context message'' that describes the goal and action semantics, and $D:S\times A \rightarrow \{0, 1\}$ denotes the binary completion indicator function.

\begin{table}[t!]
    \centering    
    \footnotesize
    \begin{tabular}{p{4.5cm}l}
    \toprule
    States (avg. per game)           &   2463.5  \\
    Action verbs (avg. per game)    &   7.4  \\
    Object types (avg. per game)    &   5.5  \\
    Object instances (avg. per state) &   10.4  \\
\midrule
    Total games                             &   31  \\    
    Total transitions & 76,369 \\
    \bottomrule
    \end{tabular}
    \caption{Corpus statistics of \datanameshort.}
    \vspace{-3mm}
    \label{tab:summary-statistics}    
\end{table}

\subsection{LLM-Sim Task}
We propose a prediction task, which we call \methodname (\methodnameshort), as a way of quantitatively evaluating the capacity of language models to serve as reliable simulators. The \methodnameshort task is defined as implementing a function $\mathcal{F}: C \times S \times A \rightarrow S \times \mathbb{R} \times \{0, 1\}$ as a world simulator that maps from a given context, state, and action (i.e. $c$, $s_{t}$, $a_{t}$) to the subsequent state, reward, and game completion status (i.e. $s_{t+1}, r_{t+1}, d_{t+1}$). 

In practice, the whole state transition simulator $\mathcal{F}$ should consider two types of state transitions: \act transitions and \tick transitions. For the example in Figure~\ref{fig:intro}, the \act transition is that the sink is turned on (\texttt{isOn=true}) after taking the action \textit{turn on sink}, and the \tick transition is that water fills up the cup in the sink when the sink is on. To better understand LLM's ability to model each of these transitions, we further decompose the simulator function $\mathcal{F}$ into three steps:  
\begin{equation*}
    \begin{split}
        s_{t+1}^\actshort &= \mathcal{F}_\actshort(c, s_t, a_t) \\
        s_{t+1} &= \mathcal{F}_\tickshort(c, s_{t+1}^\actshort) \\
        r_{t+1}, d_{t+1} &= \mathcal{F}_R(c, a_t, s_{t+1}) \\
    \end{split}
\end{equation*}


\begin{enumerate}
    \item \textbf{\Act transition simulator} $\mathcal{F}_{\actshort}: C \times S \times A \rightarrow S$ predicts $s_{t+1}^{\actshort}$ given $c$, $s_{t}$, and $a_t$, where $s_{t+1}^{\actshort}$ represents the direct state change caused by actions.
    \item \textbf{\Tick transition simulator} $\mathcal{F}_{\tickshort}: C \times S \rightarrow S$ predicts $s_{t+1}$ given $c$ and $s_{t+1}^{\actshort}$, where $s_{t+1}$ is the state that results after any \tick transitions.
    \item  \textbf{Game progress simulator} $\mathcal{F}_{R}: C \times S \times A \rightarrow \mathbb{R} \times \{0, 1\}$ predicts the reward $r_{t+1}$ and the game completion status $d_{t+1}$ given $c$, $s_{t+1}$, and $a_t$. 
\end{enumerate}

In our experiments, we measure the ability for LLMs to model $\mathcal{F}_{\actshort}$, $\mathcal{F}_{\tickshort}$, and $\mathcal{F}_{R}$ separately, as well as the complete $\mathcal{F}$ (i.e. in which all transitions are captured in a single step). We consider two variants of the \methodnameshort task:
{\flushleft \textbf{Full State Prediction:}} The LLM outputs the complete state. For example, when functioning as $\mathcal{F}$, given $c$, $s_t$ and $a_t$, the model generates the full game state $s_{t+1}$ alongside $r_{t+1}$ and $d_{t+1}$.
{\flushleft \textbf{State Difference Prediction:}} The LLM outputs only the difference between the input and output states. For example, when functioning as $\mathcal{F}$, given $c$, $s_t$ and $a_t$, the model generates only the difference between the current and subsequent game states, $\Delta((s_t, r_t, d_t), (s_{t+1}, r_{t+1}, d_{t+1}))$, as a way to reduce the need to generate redundant or unchanging information. We do not apply state difference prediction to the game progress simulator $\mathcal{F}_R$ as its output ($r_{t+1}$ and $d_{t+1}$) is not complex.

\subsection{Data}
\label{section:data}
To facilitate evaluation on the \methodnameshort task, we introduce a novel dataset of text game state transitions. Our dataset, \dataname (\datanameshort), consists of 76,369 transitions represented as $(c, s_t, r_t, d_t, a_t, s_{t+1}^\actshort, s_{t+1}, r_{t+1}, d_{t+1})$ tuples collected from 31 distinct text games. Additional corpus statistics are summarized in Table~\ref{tab:summary-statistics}.

{\flushleft \textbf{Data Collection:}} Our dataset is derived from the open \textsc{ByteSized32} corpus \cite{wang-etal-2023-bytesized32}, which consists of 32 human-authored text games that each simulate a different scientific or common-sense reasoning concept. We first modify each \textsc{ByteSized32} game to dump the game state $(s_t, r_t, d_t)$ as well as its intermediate state $s_{t+1}^{\actshort}$ at each time step $t$ as a \textsc{JSON} object. We hold out one game as an example and seed our dataset of transitions by first following the gold-label goal-following trajectory provided with each game. We then deterministically collect every valid transition that is at most one step away from the gold-label trajectory by querying the game for the set of valid actions at each step. 

{\flushleft \textbf{Additional Context:}} Each game also includes a context message, $c$, that provides additional information to the model. The context consists of four parts: \textit{action rules} describing the effect of each action on the game state, \textit{object rules} describing the meaning of each object property and whether they are affected by the game's underlying dynamics, \textit{scoring rules} describing how an agent earns reward and the conditions under which the game is won or lost, and one or two \textit{example transitions} (see Appendix~\ref{appendix-examples} for details) from the held-out game mentioned above. For each game we generate three versions of the context, one where the rules are written by a human expert (one of the game authors), and one where they are produced by an LLM with access to the game code, and one where no rules are provided. See Appendix~\ref{appendix-rules} for additional details.

\subsection{Evaluation}
Performance on \methodnameshort is determined by the model's prediction accuracy w.r.t. the ground truth labels over a dataset of test samples. Depending on the experimental condition, the LLM must model object properties (when simulating $\mathcal{F}_{\actshort}$, $\mathcal{F}_{\tickshort}$, or $\mathcal{F}$) and / or game progress (when simulating $\mathcal{F}_R$ or $\mathcal{F}$), defined as:

{\flushleft \textbf{Object Properties:}} a list of all objects in the game, along with each object's properties (\eg temperature, size) and relationships to other objects (\eg being within or on top of another object).

{\flushleft \textbf{Game Progress:}} the status of the agent w.r.t. the overall goal, consisting of the current accumulated reward, whether the game has terminated, and whether the overall goal has been achieved.

We note that in each case the LLM is provided with the ground truth previous state (when functions as $\mathcal{F}_{\tickshort}$ the previous state is $s_{t+1}^{\actshort}$) as well as the overall task context. 
That is to say, the LLM always performs a single-step prediction.

\begin{table}[t!]
\centering
\footnotesize
\begin{tabular}{p{0.6cm}p{1cm}|p{0.4cm}p{0.4cm}|p{0.4cm}p{0.4cm}p{0.4cm}p{0.4cm}}
\toprule

& State & \multicolumn{2}{c|}{\textbf{$\mathcal{F}$}} & \multicolumn{2}{c}{\textbf{$\mathcal{F}_\actshort$}} & \multicolumn{2}{c}{\textbf{$\mathcal{F}_\tickshort$}} \\

Rules & Change & \cellcolor{lp}Full & \cellcolor{lb}Diff & \cellcolor{lp}Full & \cellcolor{lb}Diff & \cellcolor{lp}Full & \cellcolor{lb}Diff \\
\midrule
\multirow{2}{*}{LLM} & \statechanged & \cellcolor{lp}59.0 & \cellcolor{lb}59.5 & \cellcolor{lp}76.1 & \cellcolor{lb}75.2 & \cellcolor{lp}44.1 & \cellcolor{lb}49.7 \\
& \stateunchanged & \cellcolor{lp}62.8 & \cellcolor{lb}72.2 & \cellcolor{lp}73.0 & \cellcolor{lb}89.5 & \cellcolor{lp}61.9 & \cellcolor{lb}93.8 \\
\midrule
\multirow{2}{*}{Human} & \statechanged & \cellcolor{lp}59.9 & \cellcolor{lb}51.6 & \cellcolor{lp}77.1 & \cellcolor{lb}68.4 & \cellcolor{lp}38.6 & \cellcolor{lb}22.2 \\
& \stateunchanged & \cellcolor{lp}63.5 & \cellcolor{lb}73.9 & \cellcolor{lp}77.5 & \cellcolor{lb}90.2 & \cellcolor{lp}73.8 & \cellcolor{lb}92.3 \\
\midrule
\multirow{2}{*}{No rule} & \statechanged & \cellcolor{lp}54.1 & \cellcolor{lb}52.2 & \cellcolor{lp}70.8 & \cellcolor{lb}67.7 & \cellcolor{lp}24.4 & \cellcolor{lb}22.3 \\
& \stateunchanged & \cellcolor{lp}56.6 & \cellcolor{lb}70.4 & \cellcolor{lp}65.3 & \cellcolor{lb}84.6 & \cellcolor{lp}73.0 & \cellcolor{lb}91.7 \\
\bottomrule
\end{tabular}
\vspace{-2mm}
\caption{\footnotesize Average accuracy per game of GPT-4 predicting the whole state transitions ($\mathcal{F}$) as well as \act transitions ($\mathcal{F}_\actshort$) and \tick transitions ($\mathcal{F}_\tickshort$). We report settings that use LLM generated rules, human written rules, or no rules. Dynamic and static denote whether the game object properties and game progress should be changed; Full and diff denote whether the prediction outcome is the full game state or state differences. Numbers are shown in percentage.}
\vspace{-1mm}
\label{tab:results-merged-gpt4}
\end{table}

\begin{table}[t!]
    \centering
    \footnotesize
    \begin{tabular}{lc}
    \toprule
    Rules & \textbf{Game Progress}\\
    \midrule
    LLM& 92.1  \\
                            Human &     81.8 \\
                            No rule &     61.5 \\

    \bottomrule
    \end{tabular}    
    \vspace{-2mm}
    \caption{\footnotesize GPT-4 game progress prediction results}
    \vspace{-4mm}
    \label{tab:results-score-gpt4}

\end{table}





\begin{table}[t!]  
    \centering  
    \footnotesize  
    \begin{tabular}{lcc}  
    \toprule  
    Game & Avg. Annotator & GPT-4\\  
    \midrule  
    bath-tub-water-temperature & 0.99 & 0.60\\  
    clean-energy & 0.50 & 0.35\\  
    take-photo & 0.83 & 0.00\\  
    metal-detector & 0.86 & 0.50\\  
    mix-paint & 0.85 & 0.50\\  
    \midrule  
    Average & 0.80 & 0.49\\  
    \bottomrule  
    \end{tabular}      
    \vspace{-1mm}
    \caption{\footnotesize Comparison between accuracy of human annotators and GPT-4 on a subset of the \datanameshort dataset. Transitions were sampled to normalize GPT-4 performance at 50\% (if possible) and annotators were tasked with modeling the complete transition function $\mathcal{F}$ and outputting the full state.} 
    \vspace{-5mm}
    \label{tab:results-human}        
\end{table}

\section{Experiments}
Figure~\ref{fig:intro} demonstrates how we evaluate the performance of a model on the \methodnameshort task using in-context learning. We evaluate the accuracy of GPT-4 in both the \textit{Full State} and \textit{State Difference} prediction regimes. The model receives the 
previous state (encoded as a \textsc{JSON} object), previous action, and context message, it 
produces the subsequent state (either as a complete \textsc{JSON} object or as a diff). See Appendix~\ref{appendix-model-details} for details.

We note that the transition dynamics between states depend primarily on the verb used in the action (\eg \textit{take}, \textit{put}, \textit{cook}, ...). In addition, some state-action pairs do not result in any changes to object properties or game progress. To ensure balance across these conditions (and increase the tractability of our experiments), we sub-sample a dataset $\mathcal{D}$ from the full \datanameshort set. Formally, let $s_{\text{in}}$ be the input state of a simulator function and $s_{\text{out}}$ be the output state of the simulator function (e.g. $s_{\text{in}}=s_t$ and $s_{\text{out}}=s_{t+1}^{\actshort}$ for $\mathcal{F}_{\actshort}$). We call any transition in which $s_{\text{out}} = s_{\text{in}}$ (according to the ground-truth) \textbf{\stateunchanged} and call each other transition \textbf{\statechanged}. Note that the \tick transition following a \statechanged \act transition is not necessarily \statechanged. For example, a state in which the agent takes an apple while the remaining objects in the environment remain the same is a \statechanged \act transition and a \stateunchanged \tick transition. We construct $\mathcal{D}$ by randomly sampling 10 \statechanged transitions and 10 \stateunchanged transitions from \datanameshort for each possible action verb (taking as many as possible if fewer than 10 exist) w.r.t \textit{\act transitions}. The resulting experimental dataset consists of 2954 transition tuples.





\section{Results}

\begin{figure*}[t!]
\centering
\includegraphics[width=\textwidth]{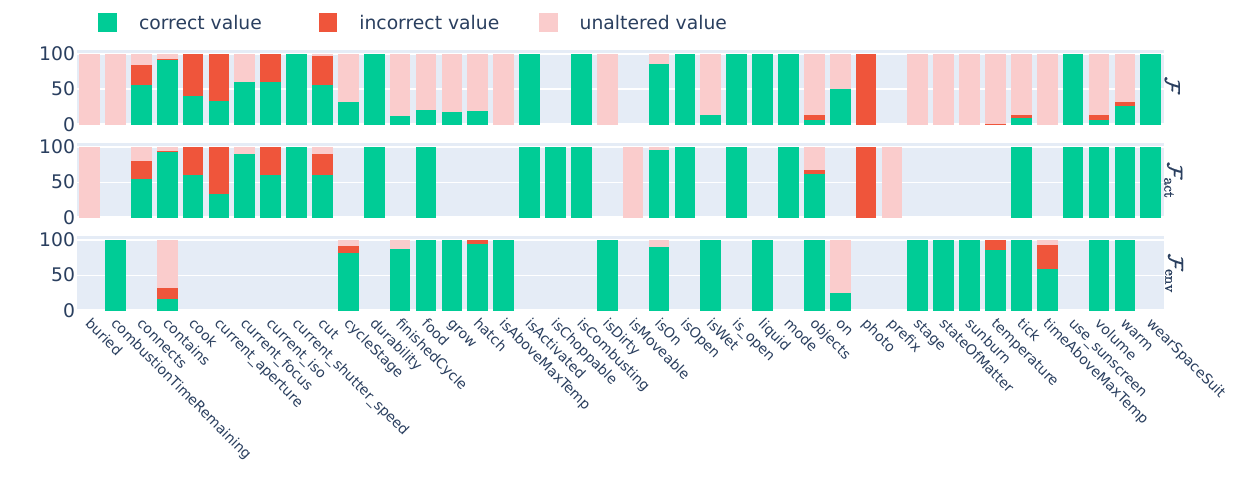}
    \vspace{-3em}
    \caption{Simulation performance of whole state transition (top), \act transitions (middle) and \tick transitions (bottom) as a function of the property being modified, in the \textit{GPT-4, full state prediction, with human written rules} condition. The \textit{x-axis} represents specific object properties, and \textit{y-axis} represents performance (0-100\%). Errors are broken down into incorrect value and unaltered value. Refer to Table~\ref{tab: property description} for the meaning of each property.}
    \label{fig:histo}
    \vspace{-1.5em}
\end{figure*}

Table~\ref{tab:results-merged-gpt4} presents the accuracy of GPT-4 simulating the whole state transitions as well as its accuracy of simulating \act transitions and \tick transitions alone.\footnote{See Appendix~\ref{appendix-gpt35results} for the results of GPT-3.5.}
We report some major observations below:

{\flushleft{\textbf{Predicting \act transitions is easier than predicting \tick transitions:}}} At best, GPT-4 is able to simulate 77.1\% of \statechanged \act transitions correctly. In contrast, GPT-4 simulates at most 49.7\% of \statechanged \tick transitions correctly. This indicates that the most challenging part of the \methodnameshort task is likely simulating the underlying environmental dynamics.

{\flushleft{\textbf{Predicting static transitions is easier than dynamic transitions:}}} 
Unsurprisingly, modeling a \stateunchanged transition is substantially easier than a \statechanged transition across most conditions. While the LLM needs to determine whether a given initial state and action will result in a state change in either case, \statechanged transitions \textit{also} require simulating the dynamics in exactly the same way as the underlying game engine by leveraging the information in the context message.

{\flushleft{\textbf{Predicting full game states is easier for dynamic states, whereas predicting state difference is easier for static states:}}} Predicting the state difference for dynamic state significantly improves the performance (>10\%) of simulating \stateunchanged transitions, while decreases the performance when simulating \statechanged transitions. This may be because state difference prediction is aimed at reducing potential format errors. However, GPT-4 is able to get the response format correct in most cases, while introducing the state difference increases the complexity of the output format of the task.

{\flushleft{\textbf{Game rules matter, and LLMs are able to generate good enough game rules:}}} Performance of GPT-4 on all three simulation tasks drops in most conditions when game rules are not provided in the context message. However, we fail to find obvious performance differences between game rules generated by human experts and by LLMs themselves.


{\flushleft{\textbf{GPT-4 can predict game progress in most cases:}}} Table~\ref{tab:results-score-gpt4} presents the results of GPT-4 predicting game progress. With game rules information in the context, GPT-4 can predict the game progress correctly in 92.1\% test cases. The presence of these rules in context is crucial: without them, GPT-4's prediction accuracy drops to 61.5\%.

{\flushleft \textbf{Humans outperform GPT-4 on the \methodnameshort task:}} 
We provide a preliminary human study on the \methodnameshort task. In particular, we take the 5 games from the \datanameshort dataset in which GPT-4 produced the worst accuracy at modeling $\mathcal{F}_\actshort$. For each game, we randomly sample 20 games with the aim of having 10 transitions where GPT-4 succeeded and 10 transitions where GPT-4 failed (note that this is not always possible because on some games GPT-4 fails/succeeds on most transitions). In addition, we balance each set of 10 transitions to have 5 \statechanged transitions and 5 \stateunchanged transitions. We instruct four human annotators (4 authors of this paper) to model as $\mathcal{F}_\actshort$ using the human-generated rules as context in a full game state prediction setting. Results are reported in Table~\ref{tab:results-human}. The overall human accuracy is 80\%, compared to the sampled LLM accuracy of 50\%, and the variation among annotators is small. This suggests that while our task is generally straightforward and relatively easy for humans, there is still a significant room for improvement for LLMs.


{\flushleft{\textbf{GPT-4 is more likely to make an error when arithmetic, common-sense, or scientific knowledge is needed:}}} Because most errors occur in modeling \statechanged transitions, we conduct an additional analysis to better understand failure modes. We use the setting with the best performance on \statechanged transitions (GPT-4, Human-written context, full state prediction) and further break down the results according to the specific object properties that are changed during the transition. Figure~\ref{fig:histo} shows, for the whole state transitions, \act transitions, and \tick transitions, the proportion of predictions that are either correct, set the property to an incorrect value, or fail to change the property value (empty columns means the property is not changed in its corresponding condition). We observe that 
 GPT-4 is able to handle most simple boolean value properties well. The errors are concentrated on non-trivial properties that requires arithmetic (e.g., \texttt{temperature}, \texttt{timeAboveMaxTemp}), common-sense (e.g., \texttt{current\_aperture}, \texttt{current\_focus}), or scientific knowledge (e.g., \texttt{on}). We also observe that when predicting the \act and \tick transitions in a single step, GPT-4 tends to focus more on \act transitions, resulting in more unaltered value errors on states that it can predict correctly when solely simulating \tick transitions.

\section{Conclusion}
We propose \dataname, a benchmark of 76,369 virtual text environment state transitions for testing LLMs as simulators. We evaluate GPT-4 on this world modeling task. Across models and conditions, the best recorded performance is 59.9\% on accurately simulating state transitions that involve non-trivial changes. 
Because simulation errors accumulate across steps, a simulator with modest single-step accuracy has limited utility in practice -- for example, after 10 steps, average simulation accuracy would reduce to $0.599^{10}$, or less than 1\%. 
Our results indicate that \textbf{LLMs are not yet able to reliably act as text world simulators}. Further error analysis shows that while LLMs are better at simulating the results of user actions, it is difficult for LLMs to handle \tick transitions and transitions that require arithmetic, common sense, or scientific knowledge. 


\section{Limitations and Ethical Concerns}


\subsection{Limitations}

This work considers two strong in-context learning LLMs, GPT-3.5 and GPT-4, in their ability to act as explicit formal simulators.
We adopt these models because they are generally the most performant off-the-shelf models across a variety of benchmarks. While we observe that even GPT-3.5 and GPT-4 achieve a modest score at the proposed task, we acknowledge that we did not exhaustively evaluate a large selection of large language models, and other models may perform better.  We provide this work as a benchmark to evaluate the performance of existing and future models on the task of accurately simulating state space transitions.

In this work, we propose two representational formalisms for representing state spaces, one that includes full state space, while the other focuses on state difference, both represented using \text{JSON} objects.  We have chosen these representations based on their popularity and compatibility with the input and output formats of most LLM pretraining data \cite[e.g.][]{fakhoury2023towards}, as well as being able to directly compare against gold standard simulator output for evaluation, though it is possible that other representational formats may be more performant at the simulation task. 

Finally, the state spaces produced in this work are focused around the domain of common-sense and early (elementary) scientific reasoning.  These tasks, such as opening containers or activating devices, were chosen because the results of these actions are common knowledge, and models are likely to be most performant in simulating these actions.  While this work does address a selection of less frequent actions and properties, it does not address using LLMs as simulators for highly domain-specific areas, such as physical or medical simulation.  A long term goal of this work is to facilitate using language models as simulators for high-impact domains, and we view this work as a stepping-stone to developing progressively more capable language model simulators.

\subsection{Ethical Concerns}
We do not foresee an immediate ethical or societal impact resulting from our work.
However, we acknowledge that as an LLM application, the proposed \methodnameshort task could be affected in some way by misinformation and hallucinations introduced by the specific LLM selected by the user. Our work highlights the issue with using LLMs as text-based world simulators. In downstream tasks, such as game simulation, LLMs may generate misleading or non-factual information. For example, if the simulator suggests burning a house to boil water, our work does not prevent this, nor do we evaluate the ethical implications of such potentially dangerous suggestions. As a result, we believe such applications are neither suitable nor safe to be deployed to a setting where they directly interact with humans, especially children, \eg in an educational setting. 
We urge researchers and practitioners to use our proposed task and dataset in a mindful manner. 

\section*{Acknowledgements}
We wish to thank the three anonymous reviewers for their helpful comments on an earlier draft of this paper.
\bibliography{anthology,custom}
\bibliographystyle{acl_natbib}

\clearpage
\appendix

\section{Model details}
\label{appendix-model-details}
For the GPT-3.5 model, we use the \code{gpt-3.5-turbo-0125} model. For the GPT-4 model, we use the \code{gpt-4-0125-preview} model. For both models, the temperature is set to 0 to get deterministic results. We also turn on the \textsc{JSON} mode of both models, which ensures that the model gives a valid \textsc{JSON} response. Our experiments cost approximately \$5,000 for OpenAI API usage.

\section{Game transition examples}
\label{appendix-examples}
We manually pick the wash-clothes game in \textsc{ByteSized32} as the example game as it contains both state transitions driven by actions and game's underlying dynamics. In tasks where the model predicts action transition, \tick transitions, or the game progress alone, we provide one corresponding in-context example. In the task that requires the model to predict everything, we offer two in-context examples in the prompt. The two examples are manually picked such that in one example the game state is changed directly by the action taken while in the other example the game state is changed by the game's underlying dynamics. 

\section{Game rules generation}
\label{appendix-rules}
\subsection{LLM generated rules}
For LLM generated rules, we manually check all of them to avoid misinformation and offensive content.

We prompt GPT-4 (\code{gpt-4-0125-preview}) with the code of each object class to acquire the rules of each object. We also provide one in-context example. We ask GPT-4 to describe the meaning of each critical property (i.e. properties that do not inherit from parent) of the object and the tick function of the object (i.e. a function that defines how object properties may change at each time step regardless of the action taken). Below is an example of our prompt of object rule generation:
\begin{tcolorbox}[fonttitle=\small\fon{pbk}\bfseries,
fontupper=\scriptsize\sffamily,
fontlower=\fon{put},
enhanced,
left=2pt, right=2pt, top=2pt, bottom=2pt,
title=Object Rule Generation Prompt]
\begin{lstlisting}[language=textgame]
You will be given a Python class which defines an object in a text game. List the classes inherited by this class and explain the properties of the object based on your understanding of the code. The properties you need to explain are commented as critical properties in the init function. If the class contains a tick method function, you should also decribe how the object properties will be changed at each game tick. Otherwise, do not explain any property. Your response should follow the format of the example below:
Here is the code for the example:
{OBJECT_CLASS_CODE}
The expected output is:
Object: Stove
Inherits: Container, Device
Properties:
maxTemperature: the maximum temperature of the stove in degrees Celsius
tempIncreasePerTick: the temperature increases per tick for objects on the stove if the stove is on.

Now here is another object class that needs you to explain:
{OBJECT_CLASS_CODE}
\end{lstlisting}
\end{tcolorbox}

For action rules generation, we prompt GPT-4 (\code{gpt-4-0125-preview}) with the code of the whole game, but unlike object rules, we do not offer any in-context example. We ask GPT-4 to describe each of the actions in the game. Below is an example of our prompt for action rule generation:
\begin{tcolorbox}[fonttitle=\small\fon{pbk}\bfseries,
fontupper=\scriptsize\sffamily,
fontlower=\fon{put},
enhanced,
left=2pt, right=2pt, top=2pt, bottom=2pt,
title=Action Rule Generation Prompt]
\begin{lstlisting}[language=textgame]
You will be given a Python program which defines an a text game. Describe the all actions based on your understanding of the code.
You can find all actions listed in the comments at the beginning of the program. You should describe all constraints of each action and how game states will be changed by taking each action.
Here is the code of the game:
{GAME_CODE}
\end{lstlisting}
\end{tcolorbox}

Similar to action rules, we generate score rules by prompting GPT-4 (\code{gpt-4-0125-preview}) with the code of the game and ask GPT-4 to describe how the game can be won or lose and how rewards can be earned. Below is an example of our prompt for score rule generation:
\begin{tcolorbox}[fonttitle=\small\fon{pbk}\bfseries,
fontupper=\scriptsize\sffamily,
fontlower=\fon{put},
enhanced,
left=2pt, right=2pt, top=2pt, bottom=2pt,
title=Score Rule Generation Prompt]
\begin{lstlisting}[language=textgame]
You will be given a Python program which defines an a text game. Describe how the game can be won or lose, and how game scores can be earned based on your understanding of the calculateScore function in the TextGame class.
Here is the code of the game. Do not describe the main function.
{GAME_CODE}
\end{lstlisting}
\end{tcolorbox}

\subsection{Human-Written Action Rules}
The action rules describe how each action can change the game states. The expert annotator reads the game description and source code for each game. They went through the list of available actions in the game and their corresponding functions in the game. Each action rule has three main parts: Action, Description, and Rules. The Action specifies the name of the action (\eg action). The Description explains the general purpose of the action (\eg connect two objects with input terminals). The Rules is an unordered list of rule descriptions that describe the constraints of the action when interacting with different objects (\eg At least one of the objects should be a wire or a multimeter) or how the rule might function under different conditions (\eg Disconnect terminal if the terminal is already connected to other objects). To ensure accuracy, the annotator plays through the game and checks if the written object rules were correctly reflected in the gameplay.

\subsection{Human-Written Object Rules}
The object rules describe the meaning of each object property (\eg temperature, size, weight, etc.) and how they will be changed at each time step. The expert annotators read the game description and source code for each game. They went through the object classes in the code script and wrote the object rules. Each object rule has three main parts: Object, Description, and Properties. The Object specifies the name of the object. The Description explains the general purpose of the object (\eg GarbageCan is a container that can hold garbage). In the Description, the inheritance of the object class has been noted. The Properties is an unordered list of property descriptions that describe each property of that object (\eg A Mold has its shape.) and their default value (\eg By default, a GameObject is not combustible.) if the object is an abstract class. For objects with tick function, there is another property describing how an object may change under each tick. To ensure accuracy, the annotator plays through the game and checks if the written object rules were correctly reflected in the gameplay.

\subsection{Human-Written Score Rules}
Score rules describe the conditions to win or lose the game and how rewards can be earned. An expert annotator (one of the \textsc{ByteSized32} game authors) creates the rules by reading the game description and the code of the score function.

\section{Prompts}
The prompts introduced in this section includes game rules that can either be human written rules or LLM generated rules. For experiments without game rules, we simply remove the rules from the corresponding prompts.

\subsection{Prompt Example: $\mathcal{F}_{\actshort}$}

\subsubsection{Full State Prediction}
\vspace{1em}

\begin{tcolorbox}[fonttitle=\small\fon{pbk}\bfseries,
fontupper=\scriptsize\sffamily,
fontlower=\fon{put},
enhanced,
left=2pt, right=2pt, top=2pt, bottom=2pt,
title=Full State Prediction Prompt ($\mathcal{F}_{\actshort}$)]
\begin{lstlisting}[language=textgame]
You are a simulator of a text game. Read the task description of a text game. Given the current game state in JSON, you need to decide the new game state after taking an action.
Your response should be in the same JSON format as the given game state.
Here is an example:
Example game task description:
Your task is to wash the dirty dishes.
Here are the descriptions of all game objects properties in the example game:
{OBJECT_RULES}
Here are the descriptions of all game actions in the example game:
{ACTION_RULES}
Here is the game state:
{GAME_STATE}
The action to take is put plate (ID: 5) in dirty cup (ID: 4)
The expected response is:
{GAME_STATE}
Here is the game that you need to simulate:
Task Description:
Your task is to figure out the weight of the cube. Use the answer action to give your answer.
Here are the descriptions of all game objects properties:
{OBJECT_RULES}
Here are the descriptions of all game actions:
{ACTION_RULES}
Here is the game state:
{GAME_STATE}
The action to take is:
look
\end{lstlisting}
\end{tcolorbox}

\subsubsection{State Difference Prediction}
\vspace{1em}

\begin{tcolorbox}[fonttitle=\small\fon{pbk}\bfseries,
fontupper=\scriptsize\sffamily,
fontlower=\fon{put},
enhanced,
left=2pt, right=2pt, top=2pt, bottom=2pt,
title=State Difference Prediction Prompt ($\mathcal{F}_{\actshort}$)]
\begin{lstlisting}[language=textgame]
You are a simulator of a text game. Read the task description of a text game. Given the current game state in JSON, you need to decide the new game state after taking an action.
Your response should be in the JSON format. It should have two keys: 'modified' and 'removed'. The 'modified' key stores a list of all the object states that are added or changed after taking the action. Keep it an empty list if no object is added or modified. The 'removed' key stores a list of uuids of the objects that are removed. Keep it an empty list if no object is removed.
Here is an example:
Example game task description:
Your task is to wash the dirty dishes.
Here are the descriptions of all game objects properties in the example game:
{OBJECT_RULES}
Here are the descriptions of all game actions in the example game:
{ACTION_RULES}
Here is the game state:
{GAME_STATE}
The action to take is put plate (ID: 5) in dirty cup (ID: 4)
The expected response is:
{GAME_STATE_DIFFERENCE}
Here is the game that you need to simulate:
Task Description:
Your task is to figure out the weight of the cube. Use the answer action to give your answer.
Here are the descriptions of all game objects properties:
{OBJECT_RULES}
Here are the descriptions of all game actions:
{ACTION_RULES}
Here is the game state:
{GAME_STATE}
The action to take is:
look
\end{lstlisting}
\end{tcolorbox}
\vfill

\clearpage
\subsection{Prompt Example: $\mathcal{F}_{\tickshort}$}

\subsubsection{Full State Prediction}
\vspace{1em}

\begin{tcolorbox}[fonttitle=\small\fon{pbk}\bfseries,
fontupper=\scriptsize\sffamily,
fontlower=\fon{put},
enhanced,
left=2pt, right=2pt, top=2pt, bottom=2pt,
title=Full State Prediction Prompt ($\mathcal{F}_{\tickshort}$)]
\begin{lstlisting}[language=textgame]
You are a simulator of a text game. Read the task description. Given the current game state in JSON, you need to decide how the game state changes in the next time step (without considering the agent actions). Rules for such changes are described as the tick function of each object.
Your response should be in the same JSON format as the given game state.
Here is an example:
Example game task description:
Your task is to wash the dirty dishes.
Here are the descriptions of all game objects properties in the example game:
{OBJECT_RULES}
Here is the game state:
{GAME_STATE}
The expected response is:
{GAME_STATE}
Here is the game that you need to simulate:
Task Description:
Your task is to figure out the weight of the cube. Use the answer action to give your answer.
Here are the descriptions of all game objects properties:
{OBJECT_RULES}
Here is the game state:
{GAME_STATE}
\end{lstlisting}
\end{tcolorbox}

\subsubsection{State Difference Prediction}
\vspace{1em}

\begin{tcolorbox}[fonttitle=\small\fon{pbk}\bfseries,
fontupper=\scriptsize\sffamily,
fontlower=\fon{put},
enhanced,
left=2pt, right=2pt, top=2pt, bottom=2pt,
title=State Difference Prediction Prompt ($\mathcal{F}_{\tickshort}$)]
\begin{lstlisting}[language=textgame]
You are a simulator of a text game. Read the task description. Given the current game state in JSON, you need to decide how the game state changes in the next time step (without considering the agent actions). Rules for such changes are described as the tick function of each object.
Your response should be in the JSON format. It should have two keys: 'modified' and 'removed'. The 'modified' key stores a list of all the object states that are added or changed after taking the action. Keep it an empty list if no object is added or modified. The 'removed' key stores a list of uuids of the objects that are removed. Keep it an empty list if no object is removed.
Here is an example:
Example game task description:
Your task is to wash the dirty dishes.
Here are the descriptions of all game objects properties in the example game:
{OBJECT_RULES}
Here is the game state:
{GAME_STATE}
The expected response is:
{GAME_STATE_DIFFERENCE}
Here is the game that you need to simulate:
Task Description:
Your task is to figure out the weight of the cube. Use the answer action to give your answer.
Here are the descriptions of all game objects properties:
{OBJECT_RULES}
Here is the game state:
{GAME_STATE}
\end{lstlisting}
\end{tcolorbox}
\vfill

\subsection{Prompt Example: $\mathcal{F}_{R}$ (Game Progress)}
\vspace{1em}

\begin{tcolorbox}[fonttitle=\small\fon{pbk}\bfseries,
fontupper=\scriptsize\sffamily,
fontlower=\fon{put},
enhanced,
left=2pt, right=2pt, top=2pt, bottom=2pt,
title=Game Progress Prediction Prompt ($\mathcal{F}_{R}$)]
\begin{lstlisting}[language=textgame]
You are a simulator of a text game. Read the task description of a text game. Given the current game state in JSON, you need to predict the current game score, whether the game is over, and whether the agent wins the game.
Your response should be a JSON with three keys: 'score', 'gameOver', and 'gameWon'. 'score' stores the current game score, 'gameOver' stores a bool value on whether the game is over, and 'gameWon' stores a bool value on whether the game is won.
Here is an example:
Example game task description:
Your task is to wash the dirty dishes.
Here are the descriptions of all game objects properties in the example game:
{OBJECT_RULES}
Here is a description of the game score function:
{SCORE_RULES}
Here is the previous game state:
{GAME_STATE}
The game score of the preivous state is:
{'score': -1, 'gameOver': False, 'gameWon': False}
The action to take is use dish soap (ID: 12) on glass (ID: 8)
{GAME_STATE}
The expected response is:
{'score': 3, 'gameOver': True, 'gameWon': True}
Here is the game that you need to simulate:
Task Description:
Your task is to figure out the weight of the cube. Use the answer action to give your answer.
Here are the descriptions of all game objects properties:
{OBJECT_RULES}
Here is a description of the game score function:
{SCORE_RULES}
Here is the previous game state:
{GAME_STATE}
The game score of the preivous state is:
{'score': 0, 'gameOver': False, 'gameWon': False}
The action to take is:
look
Here is the current game state after taking the action:
{GAME_STATE}
\end{lstlisting}
\end{tcolorbox}
\vfill

\clearpage
\subsection{Prompt Example: $\mathcal{F}$}

\subsubsection{Full State Prediction}
\vspace{1em}

\begin{tcolorbox}[fonttitle=\small\fon{pbk}\bfseries,
fontupper=\scriptsize\sffamily,
fontlower=\fon{put},
enhanced,
left=2pt, right=2pt, top=2pt, bottom=2pt,
title=Full State Prediction Prompt ($\mathcal{F}$)]
\begin{lstlisting}[language=textgame]
You are a simulator of a text game. Read the task description of a text game. Given the current game state in JSON, you need to decide the new game state after taking an action including the game score.
You may need to create new objects when you predict the new game state. You should assign the uuid of new objects starting from the UUID base given in the instructions.Your response should be in the same JSON format as the given game state.
Note that while game states can be changed by actions, some game states may change over the time, which is described in the tick function of each object class. 
Here are two examples of both cases. Both examples are from the same example game.
Example game task description:
Your task is to wash the dirty dishes.
Here are the descriptions of all game objects properties in the example game:
{OBJECT_RULES}
Here are the descriptions of all game actions in the example game:
{ACTION_RULES}
Here is a description of the score function of the example game:
{SCORE_RULES}
In the first example, the game state is changed by an action:
Here is the game state:
{GAME_STATE}
The current game UUID base is 12
The action to take is: put plate (ID: 5) in dirty cup (ID: 4)
The expected response is:
{GAME_STATE}
In the second example from the same example game, the game state is changed over the time. Note that while in this example the game state is changed by time only, it is possible that a game state is changed by both an action and time.
Here is the game state:
{GAME_STATE}
The current game UUID base is 13
The action to take is: eat dishwasher (ID: 2) with dirty plate (ID: 5)
The expected response is:
{GAME_STATE}
Here is the game that you need to simulate:
{OBJECT_RULES}
Here are the descriptions of all game actions:
{ACTION_RULES}
Here is a description of the game score function:
{SCORE_RULES}
Here is the game state:
{GAME_STATE}
The current game UUID base is 12
The action to take is:
look
\end{lstlisting}
\end{tcolorbox}
\vfill

\subsubsection{State Difference Prediction}
\vspace{1em}

\begin{tcolorbox}[fonttitle=\small\fon{pbk}\bfseries,
fontupper=\scriptsize\sffamily,
fontlower=\fon{put},
enhanced,
left=2pt, right=2pt, top=2pt, bottom=2pt,
title=State Difference Prediction Prompt ($\mathcal{F}$)]
\begin{lstlisting}[language=textgame]
You are a simulator of a text game. Read the task description and the current environment observation description. Given the current game state in \textsc{JSON}, you need to decide the new game state after taking an action.
Your response should be in the \textsc{JSON} format. It should have three keys: 'modified', 'removed', and 'score'. The 'modified' key stores a list of all the object states that are added or changed after taking the action. Keep it an empty list if no object is added or modified. The 'removed' key stores a list of uuids of the objects that are removed. Keep it an empty list if no object is removed. The 'score' key stores a dictionary with three keys: 'score' is the current game score, 'gameOver' is a boolean of whether the game is over, and 'gameWon' is a boolean of whether the agent won the game. If a player earns a score or wins/loses the game, you should reflect that change in the dictionary saved under the 'score' key. Otherwise, you should set value of the 'score' key to an empty dictionary.Note that while game states can be changed by actions, some game states may change over the time, which is described in the tick function of each object class.
Note that while game states can be changed by actions, some game states may change over the time, which is described in the tick function of each object class. 
Here are two examples of both cases. Both examples are from the same example game.
Example game task description:
Your task is to wash the dirty dishes.
Here are the descriptions of all game objects properties in the example game:
{OBJECT_RULES}
Here are descriptions of all game actions in the example game:
{ACTION_RULES}
Here is a description of the score function of the example game:
{SCORE_RULES}
In the first example, the game state is changed by an action:
Current observation: 
{GAME_OBSERVATION}
Here is the game state:
{GAME_STATE}
The action to take is put dirty plate (ID: 5) in mug (ID: 6)
The expected response is:
{GAME_STATE_DIFFERENCE}
In the second example from the same example game, the game state is changed over the time. Note that while in this example the game state is changed by time only, it is possible that a game state is changed by both an action and time.
Current observation: 
{Example_2 observation}
Here is the game state:
{GAME_STATE}
The action to take is eat dishwasher (ID: 2) with dirty plate (ID: 5)
The expected response is:
{GAME_STATE_DIFFERENCE}
Here is the game that you need to simulate:
Task Description:
Your task is to boil water.
Here are the descriptions of all game objects properties:
{OBJECT_RULES}
Here are the descriptions of all game actions:
{ACTION_RULES}
Here is a description of the score function of the game:
{SCORE_RULES}
Current observation: 
{GAME_OBSERVATION}
Here is the game state:
{GAME_STATE}
The current game UUID base is 12
The action to take is:
look

\end{lstlisting}
\end{tcolorbox}
\vfill

\clearpage
\subsection{Other Examples}

Below is an example of the rule of an action:
\begin{tcolorbox}[fonttitle=\small\fon{pbk}\bfseries,
fontupper=\scriptsize\sffamily,
fontlower=\fon{put},
enhanced,
left=2pt, right=2pt, top=2pt, bottom=2pt,
title=Action Rule Example]
\begin{lstlisting}[language=textgame]
put:
Description: put an object into a target container
Rules:
1. The target must be a container (Container)
2. The target container must be open
3. The object must be in the inventory
4. The object must be moveable (isMoveable)
\end{lstlisting}
\end{tcolorbox}

Below is an example of the rule of an object:
\begin{tcolorbox}[fonttitle=\small\fon{pbk}\bfseries,
fontupper=\scriptsize\sffamily,
fontlower=\fon{put},
enhanced,
left=2pt, right=2pt, top=2pt, bottom=2pt,
title=Object Rule Example]
\begin{lstlisting}[language=textgame]
Object: Container
Description: Abstract class for things that can be considered 'containers' (e.g. a drawer, a box, a table, a shelf, etc.)
Properties:
- A Container is a container.
- A Container could be opened (e.g., e.g. a drawer, a door, a box, etc.), or is it always 'open' (e.g. a table, a shelf, etc.). 
- A Container has a property indicating if it is opened.
- A Container has a property indicating the prefix to use when referring to the container (e.g. "in the drawer", "on the table", etc.). By default, the prefix is 'in'
\end{lstlisting}
\end{tcolorbox}

Below is an example of the score rule:
\begin{tcolorbox}[fonttitle=\small\fon{pbk}\bfseries,
fontupper=\scriptsize\sffamily,
fontlower=\fon{put},
enhanced,
left=2pt, right=2pt, top=2pt, bottom=2pt,
title=Score Rule Example]
\begin{lstlisting}[language=textgame]
The player wins the game by getting all dishes clean.
The player gets one point for each dish that is cleaned.
The player loses one point for each dish that is made dirty.
\end{lstlisting}
\end{tcolorbox}

Below is an example of a game state:
\begin{tcolorbox}[fonttitle=\small\fon{pbk}\bfseries,
fontupper=\scriptsize\sffamily,
fontlower=\fon{put},
enhanced,
left=2pt, right=2pt, top=2pt, bottom=2pt,
title=Game State Example]
\begin{lstlisting}[language=textgame]
{'game_state': [{'name': 'agent (ID: 0)', 'uuid': 0, 'type': 'Agent', 'properties': {'isContainer': True, 'isMoveable': True, 'isOpenable': False, 'isOpen': True, 'containerPrefix': 'in'}, 'contains': ['plate (ID: 5)', 'mug (ID: 6)', 'knife (ID: 7)']}, {'name': 'plate (ID: 5)', 'uuid': 5, 'type': 'Dish', 'properties': {'isContainer': True, 'isMoveable': True, 'isOpenable': False, 'isOpen': True, 'containerPrefix': 'on', 'dishType': 'plate', 'isDirty': True, 'foodMessName': 'orange'}, 'contains': []}, {'name': 'mug (ID: 6)', 'uuid': 6, 'type': 'Dish', 'properties': {'isContainer': True, 'isMoveable': True, 'isOpenable': False, 'isOpen': True, 'containerPrefix': 'in', 'dishType': 'mug', 'isDirty': True, 'foodMessName': 'sandwhich'}, 'contains': []}, {'name': 'knife (ID: 7)', 'uuid': 7, 'type': 'Dish', 'properties': {'isContainer': True, 'isMoveable': True, 'isOpenable': False, 'isOpen': True, 'containerPrefix': 'in', 'dishType': 'knife', 'isDirty': True, 'foodMessName': 'apple (ID: 11)'}, 'contains': []}, {'name': 'dishwasher (ID: 2)', 'uuid': 2, 'type': 'DishWasher', 'properties': {'isContainer': True, 'isMoveable': False, 'isOpenable': True, 'isOpen': True, 'containerPrefix': 'in', 'isDevice': True, 'isActivatable': True, 'isOn': False, 'cycleStage': 0, 'finishedCycle': False}, 'contains': ['cup (ID: 4)']}, {'name': 'cup (ID: 4)', 'uuid': 4, 'type': 'Dish', 'properties': {'isContainer': True, 'isMoveable': True, 'isOpenable': False, 'isOpen': True, 'containerPrefix': 'in', 'dishType': 'cup', 'isDirty': True, 'foodMessName': 'peanut butter'}, 'contains': []}, {'name': 'bottle of dish soap (ID: 3)', 'uuid': 3, 'type': 'DishSoapBottle', 'properties': {'isContainer': False, 'isMoveable': True, 'isDevice': True, 'isActivatable': True, 'isOn': False}, 'contains': []}, {'name': 'glass (ID: 8)', 'uuid': 8, 'type': 'Dish', 'properties': {'isContainer': True, 'isMoveable': True, 'isOpenable': False, 'isOpen': True, 'containerPrefix': 'in', 'dishType': 'glass', 'isDirty': False}, 'contains': []}, {'name': 'bowl (ID: 9)', 'uuid': 9, 'type': 'Dish', 'properties': {'isContainer': True, 'isMoveable': True, 'isOpenable': False, 'isOpen': True, 'containerPrefix': 'in', 'dishType': 'bowl', 'isDirty': False}, 'contains': []}, {'name': 'banana (ID: 10)', 'uuid': 10, 'type': 'Food', 'properties': {'isContainer': False, 'isMoveable': True, 'isFood': True}, 'contains': []}, {'score': -1, 'gameOver': False, 'gameWon': False}]}
\end{lstlisting}
\end{tcolorbox}

Below is an example of a \textsc{JSON} that describes the difference of two game states:
\begin{tcolorbox}[fonttitle=\small\fon{pbk}\bfseries,
fontupper=\scriptsize\sffamily,
fontlower=\fon{put},
enhanced,
left=2pt, right=2pt, top=2pt, bottom=2pt,
title=Game State Difference Example]
\begin{lstlisting}[language=textgame]
{'modified': [{'name': 'agent (ID: 0)', 'uuid': 0, 'type': 'Agent', 'properties': {'isContainer': True, 'isMoveable': True, 'isOpenable': False, 'isOpen': True, 'containerPrefix': 'in'}, 'contains': ['mug (ID: 6)', 'knife (ID: 7)']}, {'name': 'mug (ID: 6)', 'uuid': 6, 'type': 'Dish', 'properties': {'isContainer': True, 'isMoveable': True, 'isOpenable': False, 'isOpen': True, 'containerPrefix': 'in', 'dishType': 'mug', 'isDirty': True, 'foodMessName': 'sandwhich'}, 'contains': ['plate (ID: 5)']}], 'removed': [], 'score': {}}
\end{lstlisting}
\end{tcolorbox}

\section{GPT-3.5 results}
\label{appendix-gpt35results}
Table~\ref{tab:results-merged-gpt35} and Table~\ref{tab:results-score-gpt35} shows the performance of a GPT-3.5 simulator predicting objects properties and game progress respectively. There is a huge gap between the GPT-4 performance and GPT-3.5 performance, providing yet another example of how fast LLM develops in the two years. It is also worth notices that the performance difference is larger when no rules is provided, indicating that GPT-3.5 is especially weak at applying common sense knowledge to this few-shot world simulation task.

\begin{table}[t!]  
    \centering  
    \footnotesize  
    \begin{tabular}{p{0.6cm}p{1cm}|p{0.4cm}p{0.4cm}|p{0.4cm}p{0.4cm}p{0.4cm}p{0.4cm}}
    \toprule  
    & State & \multicolumn{2}{c|}{\textbf{$\mathcal{F}$}} & \multicolumn{2}{c}{\textbf{$\mathcal{F}_\actshort$}} & \multicolumn{2}{c}{\textbf{$\mathcal{F}_\tickshort$}} \\
    
    
    Rules & Change & \cellcolor{lp}Full & \cellcolor{lb}Diff & \cellcolor{lp}Full & \cellcolor{lb}Diff & \cellcolor{lp}Full & \cellcolor{lb}Diff \\
    
    \midrule  
    \multirow{2}{*}{LLM} &   \statechanged   & \cellcolor{lp}34.5 & \cellcolor{lb}21.4 & \cellcolor{lp}36.0 & \cellcolor{lb}31.7 & \cellcolor{lp}7.8 & \cellcolor{lb}2.9 \\  
                         &    \stateunchanged & \cellcolor{lp}37.5 & \cellcolor{lb}54.0 & \cellcolor{lp}44.6 & \cellcolor{lb}65.9 & \cellcolor{lp}41.8 & \cellcolor{lb}63.1 \\  
    \midrule
    \multirow{2}{*}{Human} &    \statechanged   & \cellcolor{lp}26.8 & \cellcolor{lb}21.2 & \cellcolor{lp}43.3 & \cellcolor{lb}36.1 & \cellcolor{lp}12.5 & \cellcolor{lb}0.4 \\  
                         &    \stateunchanged & \cellcolor{lp}35.6 & \cellcolor{lb}58.9 & \cellcolor{lp}42.3 & \cellcolor{lb}64.7 & \cellcolor{lp}22.0 & \cellcolor{lb}74.2 \\  
    \midrule
    \multirow{2}{*}{No rule} &    \statechanged   & \cellcolor{lp}15.4 & \cellcolor{lb}23.5 & \cellcolor{lp}43.8 & \cellcolor{lb}35.7 & \cellcolor{lp}1.7 & \cellcolor{lb}0.8 \\  
                         &    \stateunchanged & \cellcolor{lp}26.9 & \cellcolor{lb}50.0 & \cellcolor{lp}35.2 & \cellcolor{lb}63.0 & \cellcolor{lp}17.2 & \cellcolor{lb}54.8 \\  
    \bottomrule  
    \end{tabular}      
    \caption{\footnotesize Average accuracy per game of GPT-3.5 predicting the whole state transitions ($\mathcal{F}$) as well as \act transitions ($\mathcal{F}_\actshort$) and \tick transitions ($\mathcal{F}_\tickshort$). We report settings that use LLM generated rules, human written rules, or no rules. Dynamic and static denote whether the game object properties and game progress should be changed; Full and diff denote whether the prediction outcome is the full game state or state differences. Numbers shown in percentage.}
    \label{tab:results-merged-gpt35}  
\end{table}  

\begin{table}[t!]
    \centering
    \footnotesize
    \begin{tabular}{lc}
    \toprule
    Rules & \textbf{Game Progress}\\
    \midrule
    LLM& 73.9  \\
                            \midrule
                            Human &     63.3 \\
                            \midrule
                            No rule &     64.2 \\

    \bottomrule
    \end{tabular}    
    \caption{\footnotesize GPT-3.5 game progress prediction results}
    \label{tab:results-score-gpt35}
\end{table}

\begin{table*}[t!]      
    \centering      
    \footnotesize      
    \begin{tabular}{ll}    
    \toprule    
    Property Name & Description \\    
    \midrule   
buried&Objects buried in the room\\
combustionTimeRemaining&Number of time steps remaining to combust of a combusting object\\
connects&Electrical objects connecting to the current object\\
contains&Objects in the current object\\  
cook&How an ingredient is cooked\\  
current\_aperture&Current aperture of a camera\\  
current\_focus&The object that the camera is currently focusing on\\  
current\_iso&Current ISO of a camera\\
current\_shutter\_speed&Current shutter speed of a camera\\
cut&How an ingredient is cut\\  
cycleStage&The current stage of the washing machine's cycle (running/washing/finished).\\
durability&Number of times left for a shovel to dig something\\
finishedCycle&A boolean indicator of whether the washing machine has finished\\
food&The food level of a young bird. Reduce 1 if the young bird is not fed at each time step.\\
grow&Number of time steps that a young bird has grown\\
hatch&Number of time steps that an egg is hatched\\
isAboveMaxTemp&Whether the temperature of the current food is above its maximum preservation temperature\\
isActivated&Whether a device is activated\\
isChoppable&Whether an object is choppable\\
isCombusting&Whether an object is combusting\\
isDirty&Whether a dish is dirty\\
isMoveable&Whether the current object is moveable\\
isOn&Whether a device is turned on\\
isOpen&Whether a container is open\\
isWet&Whether a clothes is wet\\
is\_open&Whether a door is open\\
liquid&Whether there is liquid in a container\\
mode&Mode of a multimeter\\
objects&Record of the number of time steps that each object is on the inclined plane\\  
on&Whether a light bulb is on\\  
photo&The object that the camera has taken a picture of\\  
prefix&Prefix abstract to describe the object. E.g., \textbf{a} tree and \textbf{some} firewood\\  
stage&Life stage of a bird\\
stateOfMatter&State of matter of a substance\\
sunburn&Whether the player's skin is burnt by the sun\\
temperature&Object temperature\\
tick&Number of ticks that an object is placed on an inclined plane\\
timeAboveMaxTemp&Number of time steps that a food is above its maximum preservation temperature\\
use\_sunscreen&Whether the player has used the sunscreen\\
volume&Volume of an object\\
warm&The warmth received by an egg during its hatching stage\\
wearSpaceSuit&Whether the agent wears the spacesuit\\
\bottomrule  
  
\end{tabular}  
\caption{Description of object properties mentioned in Figure~\ref{fig:histo}} 
\label{tab: property description}
\end{table*}  

\section{Histograms}
\begin{figure*}[t!]
\centering
\begin{subfigure}{\textwidth}
\caption{Human-generated rules.}
\includegraphics[width=\textwidth]{figures/imagesv2/gpt-4-0125-preview_apr24_full_hwr_action-tick-full_hist_property.pdf}
\end{subfigure}

\begin{subfigure}{\textwidth}
\caption{LLM-generated rules.}
\includegraphics[width=\textwidth]{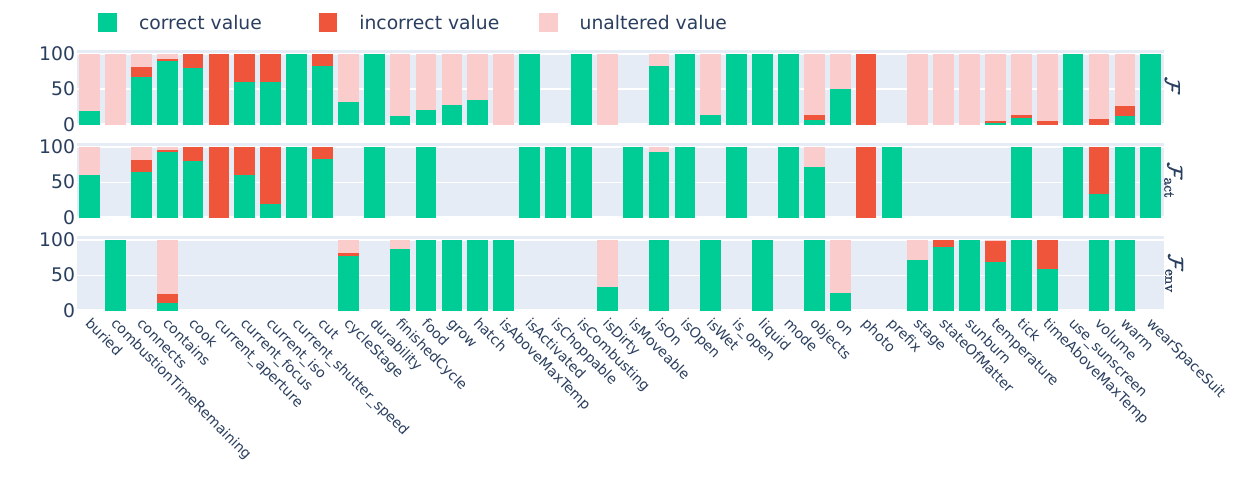}
\end{subfigure}

\begin{subfigure}{\textwidth}
\caption{No rules.}
\includegraphics[width=\textwidth]{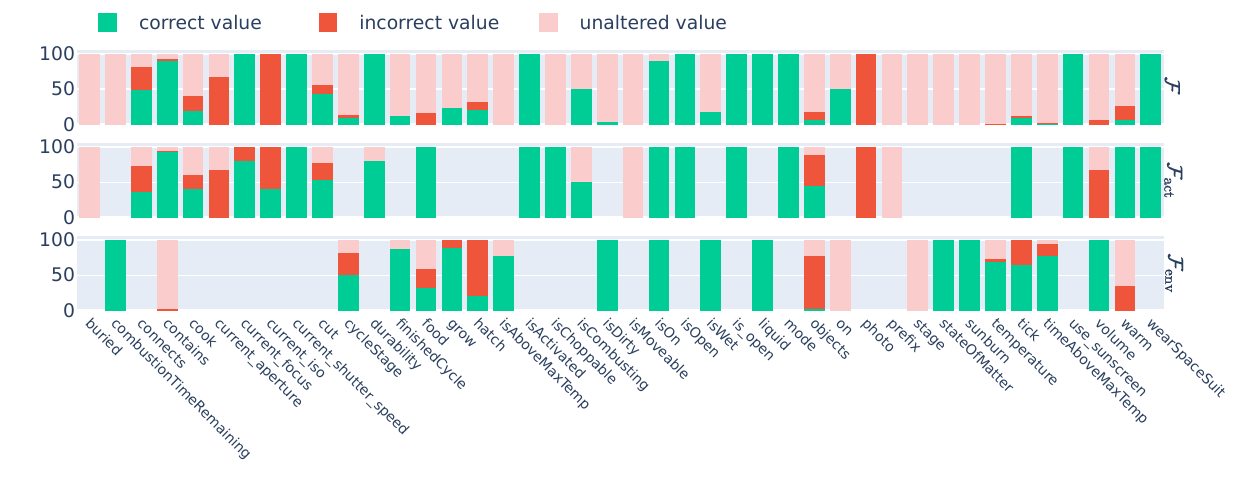}
\end{subfigure}
\caption{GPT-4 - Full State prediction from a) Human-generated rules, b) LLM-generated rules, and c) No rules.
} 
\label{appfig:gpt4-full}
\end{figure*}

\begin{figure*}[t!]
\centering
\begin{subfigure}{\textwidth}
\caption{Human-generated rules.}
\includegraphics[width=\textwidth]{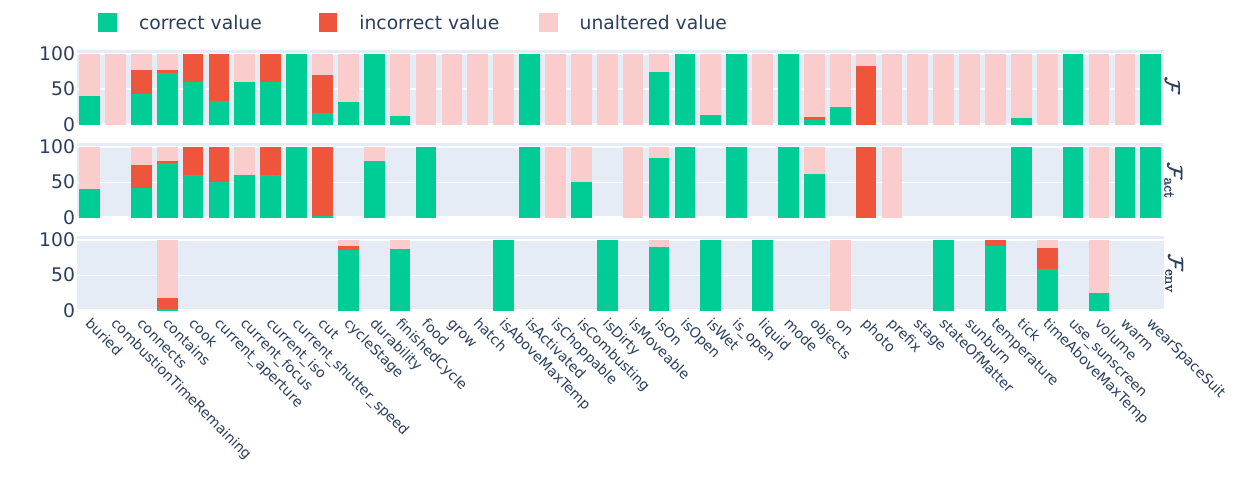}
\end{subfigure}

\begin{subfigure}{\textwidth}
\caption{LLM-generated rules.}
\includegraphics[width=\textwidth]{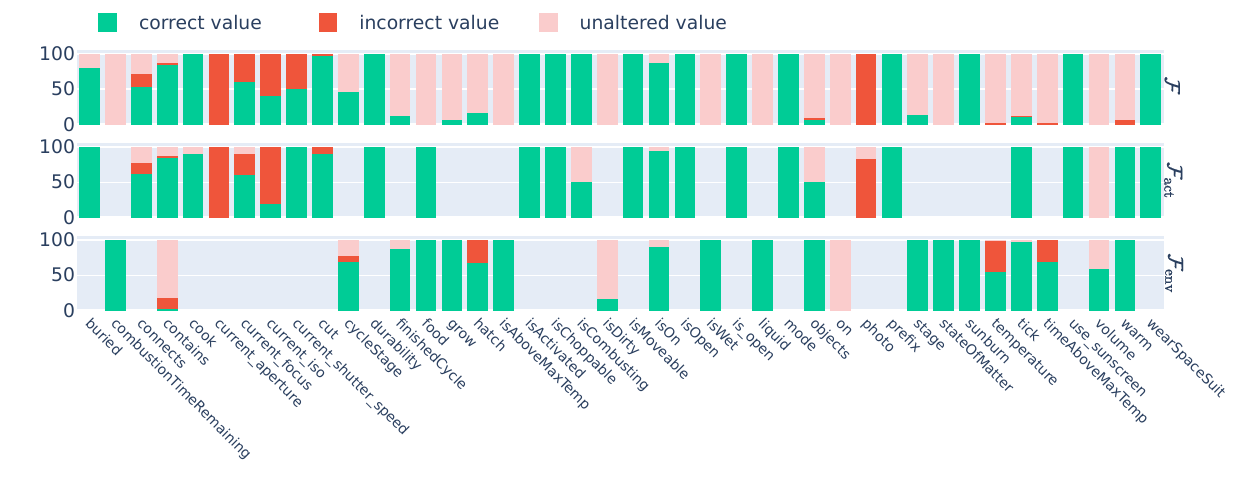}
\end{subfigure}

\begin{subfigure}{\textwidth}
\caption{No rules.}
\includegraphics[width=\textwidth]{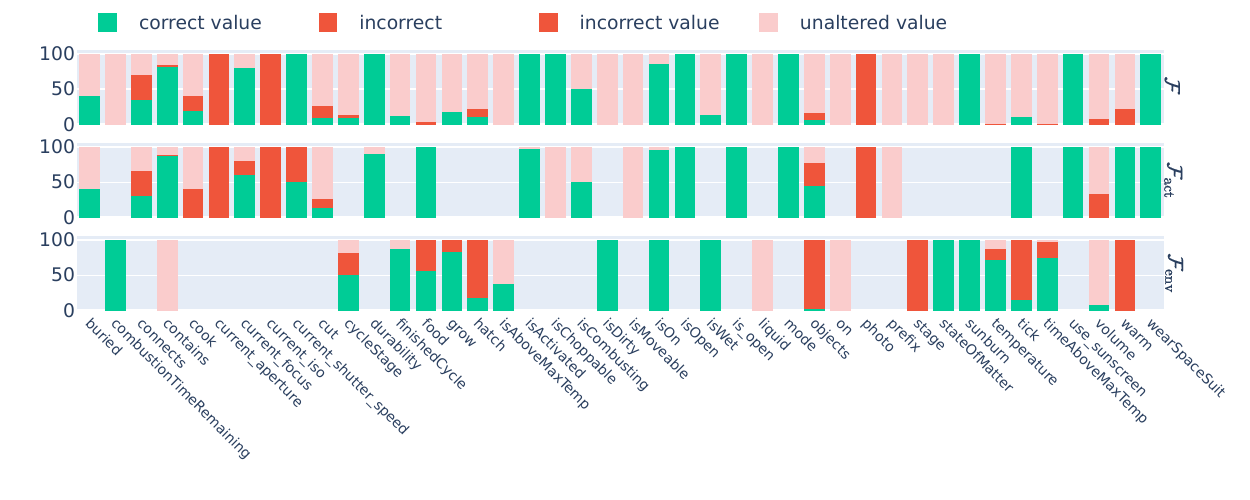}
\end{subfigure}
\caption{GPT-4 - Difference prediction from a) Human-generated rules, b) LLM-generated rules, and c) No rules.
} 
\label{appfig:gpt4-diff}
\end{figure*}

\begin{figure*}[t!]
\centering
\begin{subfigure}{\textwidth}
\caption{Human-generated rules.}
\includegraphics[width=\textwidth]{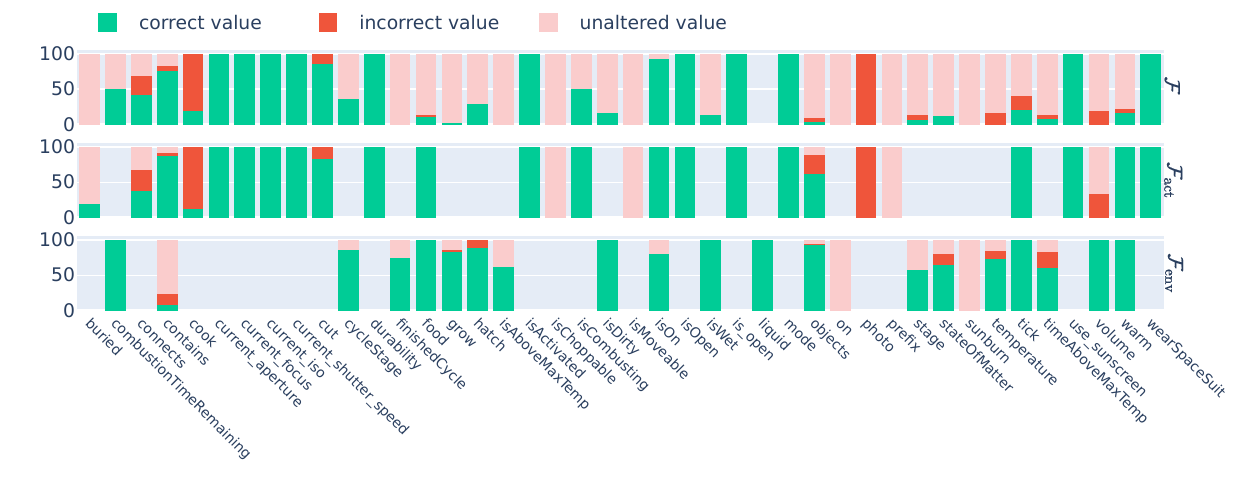}
\end{subfigure}

\begin{subfigure}{\textwidth}
\caption{LLM-generated rules.}
\includegraphics[width=\textwidth]{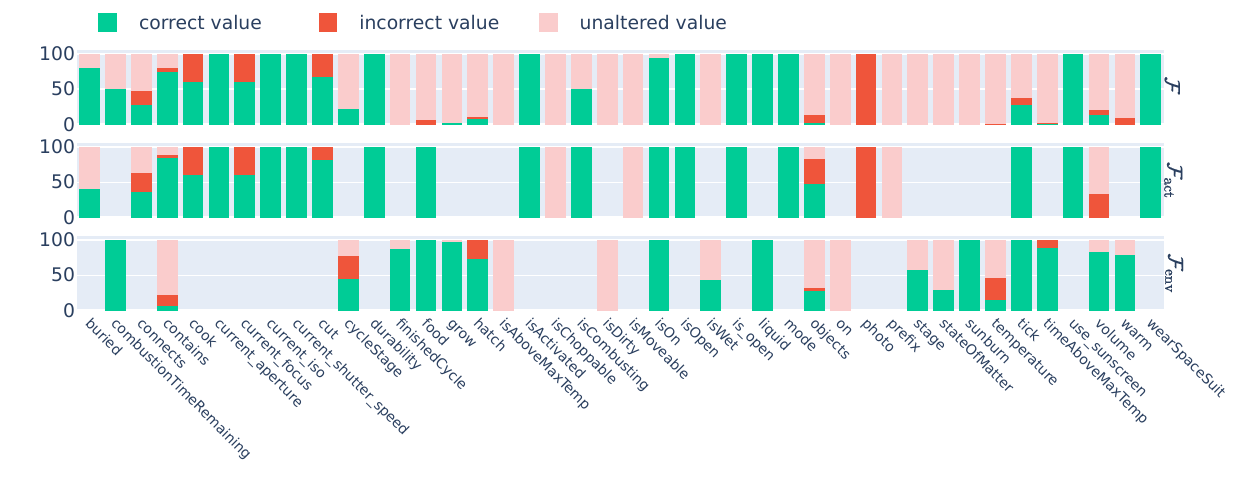}
\end{subfigure}

\begin{subfigure}{\textwidth}
\caption{No rules.}
\includegraphics[width=\textwidth]{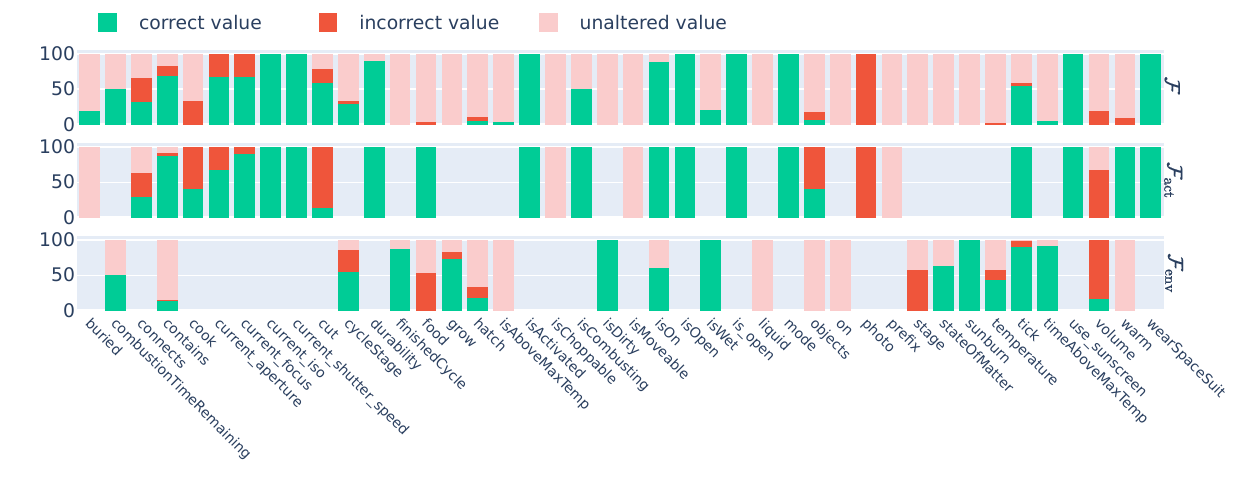}
\end{subfigure}
\caption{GPT-3.5 - Full State prediction from a) Human-generated rules, b) LLM-generated rules, and c) No rules.
} 
\label{appfig:gpt35-full}
\end{figure*}

\begin{figure*}[t!]
\centering
\begin{subfigure}{\textwidth}
\caption{Human-generated rules.}
\includegraphics[width=\textwidth]{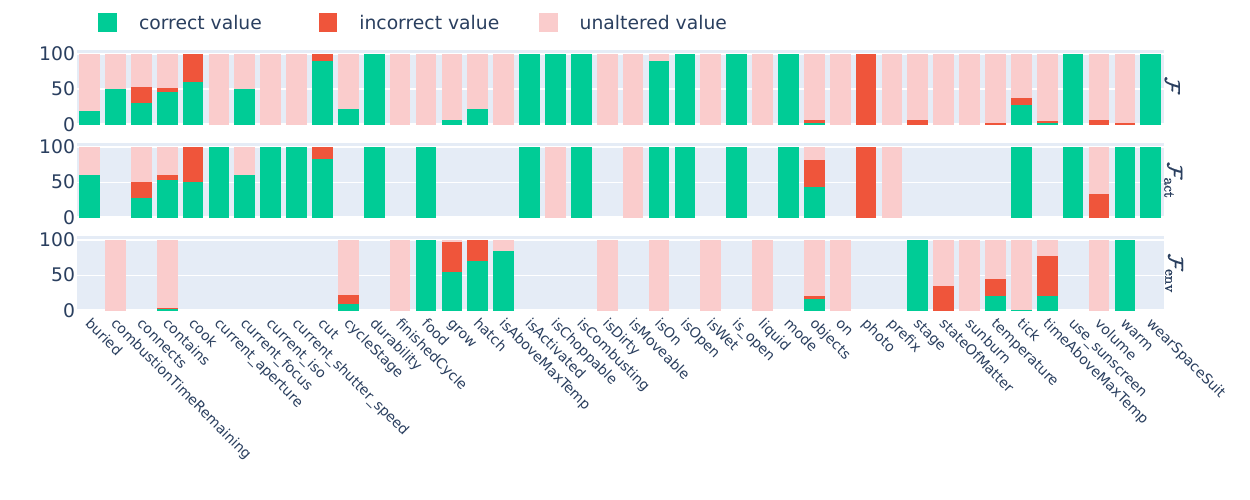}
\end{subfigure}

\begin{subfigure}{\textwidth}
\caption{LLM-generated rules.}
\includegraphics[width=\textwidth]{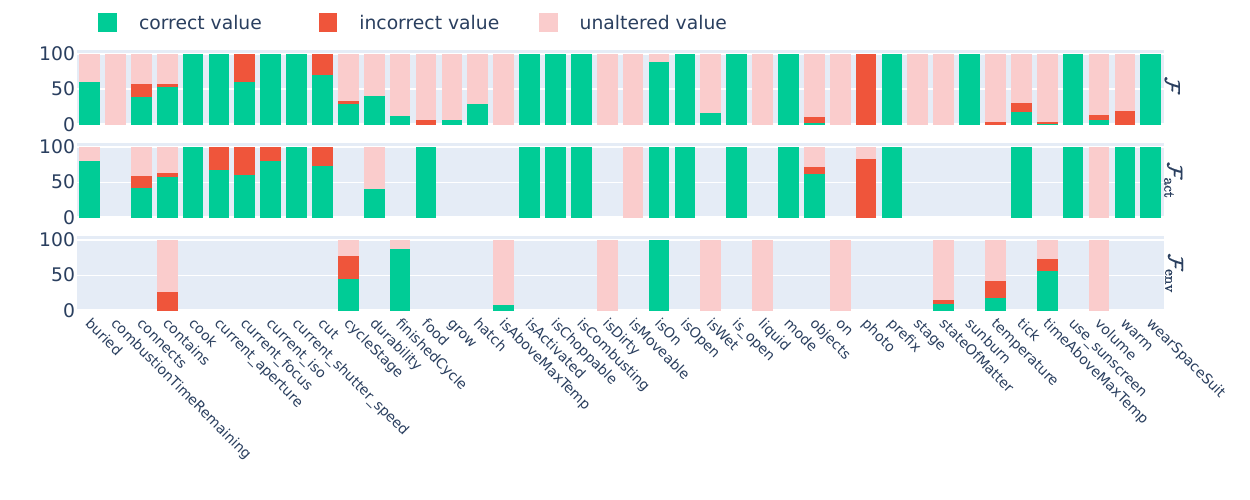}
\end{subfigure}

\begin{subfigure}{\textwidth}
\caption{No rules.}
\includegraphics[width=\textwidth]{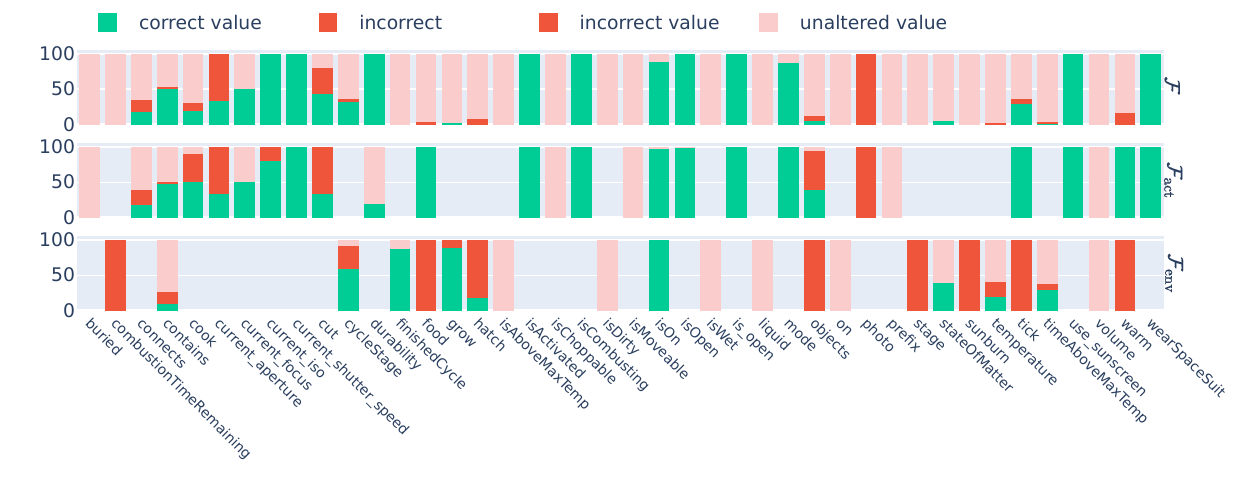}
\end{subfigure}
\caption{GPT-3.5 - Difference prediction from a) Human-generated rules, b) LLM-generated rules, and c) No rules.
} 
\label{appfig:gpt35-diff}
\end{figure*}

\begin{enumerate}
    \item In Figure~\ref{appfig:gpt4-full}, we show detailed experimental results on the \textbf{full state prediction task} performed by \textbf{GPT-4}. 
    \item In Figure~\ref{appfig:gpt4-diff}, we show detailed experimental results on the \textbf{state difference prediction task} performed by \textbf{GPT-4}. 
    \item In Figure~\ref{appfig:gpt35-full}, we show detailed experimental results on the \textbf{full state prediction task} performed by \textbf{GPT-3.5}. 
    \item In Figure~\ref{appfig:gpt35-diff}, we show detailed experimental results on the \textbf{state difference prediction task} performed by \textbf{GPT-3.5}. 
\end{enumerate}

\end{document}